# Domain Knowledge in Artificial Intelligence: Using Conceptual Modeling to Increase Machine Learning Accuracy and Explainability

**Veda C. Storey, Jeffrey Parsons, Arturo Castellanos Bueso, Monica Chiarini Tremblay, Roman Lukyanenko, Alfred Castillo, Wolfgang Maass**

**Abstract**

Machine learning enables the extraction of useful information from large, diverse datasets. However, despite many successful applications, machine learning continues to suffer from performance and transparency issues. These challenges can be partially attributed to the limited use of domain knowledge by machine learning models. This research proposes using the domain knowledge represented in conceptual models to improve the preparation of the data used to train machine learning models. We develop and demonstrate a method, called the *Conceptual Modeling for Machine Learning (CMML)*, which is comprised of guidelines for data preparation in machine learning and based on conceptual modeling constructs and principles. To assess the impact of CMML on machine learning outcomes, we first applied it to two real-world problems to evaluate its impact on model performance. We then solicited an assessment by data scientists on the applicability of the method. These results demonstrate the value of CMML for improving machine learning outcomes.

## 1 Introduction

Machine learning (ML) has been widely adopted to support a range of business functions, industries, and daily tasks, although challenges in its adoption remain. Among the important ones are understanding and preparing data for use in the ML algorithms that comprise the models. ML models are built from data so the quality of the models depends upon the quality of the data used to create them [41]. However, current ML approaches lack streamlined processes for improving the data quality and typically follow ad-hoc manual processes [69]. In traditional data management, conceptual modeling (CM) comprises approaches to understanding how real-world entities and relationships among them are represented in data, typically by representing data semantics via graphical abstractions [21]. However, the use of conceptual models to understand data is limited. We, therefore, propose using CM to improve ML by preparing data in ways that better reflect knowledge about what the data represents.

The ML community and, more broadly, the artificial intelligence (AI) community have long emphasized the need for a data-centric approach that focuses on the quality of the training data. As Press [64] notes: "In the dominant model-centric approach to AI, according to [Andrew] Ng, you hold the data fixed and iteratively improve the model until the desired results are achieved. In the nascent data-centric approach to AI, consistency of data is paramount. To get to the right results, you hold the model or code fixed and iteratively improve the quality of the data." The data-centric AI movement aims to address the lack of tooling, best practices, and infrastructure for managing data in modern ML systems [67]. The activities include data collection, data labeling, data preprocessing, data augmentation, data quality evaluation, data debt, and data governance. In practice, model-centric and data-centric approaches are iterativeley used. After training a baseline model of sufficient quality, data quality is improved by applying data-centric activities until the data quality stabilizes. This dual process is iteratively applied until a stopping rule applies.

Performing these activities well is challenging, with errors related to data management often leading to compounding events ("data cascades") that can cause negative effects ranging from low ML performance to discrimination and biases. Sambasivan et al [67] found that these data cascades are avoidable through intentional practices, such as dataset documentation of data pipelines [55].

Another major challenge is ensuring process transparency when building models [13]. Machine learning applications depend upon the experience and intuition of data scientists, which are often undocumented [6, 38]. Projects must be managed effectively by making the ML process transparent, repeatable, and auditable. For example, assessing highly sophisticated and opaque transformations of the input data can result in high accuracy when evaluating models, but provide little or no transparency on how the original features weigh on the predictions [1, 33].

The objective of this research is to *create a mechanism for improving both machine learning performance and process transparency.* We propose a method for using domain knowledge, as found in conceptual models, to augment the data preparation needed to create training data sets to build machine learning models. The method is developed primarily for structured databases, but can be extended to non-structured databases.

The contribution is the creation and evaluation of a Conceptual Modeling for Machine Learning (CMML) method, comprised of a set of guidelines that can be applied to create effective training data sets prior to building machine learning models. To evaluate the usefulness of CMML, we first apply the method to data obtained from two real-world organizations in the United States to assess its potential to improve machine learning performance. We then conduct an applicability check of the proposed guidelines by engaging data scientists in a focus group setting.

Section 2 of this paper provides an overview of machine learning and conceptual modeling. Section 3 presents the Conceptual Modeling for Machine Learning (CMML) method. Section 4 reports on the evaluation tasks, with the implications discussed in Section 5. Section 6 concludes the paper.

## 2 Machine Learning and Conceptual Modeling

This section briefly reviews concepts related to machine learning needed to understand the development of the CMML method.

### 2.1 Supervised Machine Learning

The most common type of machine learning is no doubtly supervised ML, although the intent of our work is to be generalizable to other types, such as unsupervised or reinforcement learning. This is because we focus on data semantics, with data a required input of all types of ML. A learning machine is a computer program that can improve its performance with experience for some class of tasks and performance measures [11]. Supervised machine learning uses labeled training and testing data[1] to build and evaluate models that represent patterns inferred from the data. Training data comprises variables (*features*), which are used in raw form or transformed into new features, to predict a target attribute of interest. Features sit between the data and models in the machine learning pipeline [25, 85].

Increasingly, the goal of machine learning is to improve performance while enacting fair, transparent, auditable, and repeatable processes [4]. Much of the effort in deploying machine learning algorithms goes into the design of preprocessing pipelines and data transformations that result in a representation of the data that can support effective machine learning [9].

For supervised machine learning, performance is the ability of a machine learning model to predict data in accordance with data used as input [11] as shown below.

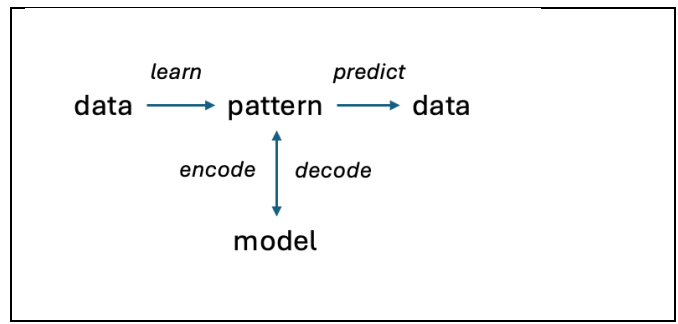

Supervised learning guides the learner in acquiring knowledge in a domain through examples so that new cases can be handled appropriately based on the implicitly learned patterns derived from similar cases [11]. The accuracy of a model measures performance in predicting values of interest for unseen instances (i.e., ones not used to build the model).

The performance of machine learning performance is generally affected by: fit between data and model complexity (for instance, more complex methods, such as deep learning neural networks yielding more abstract, and ultimately more useful, representations for large datasets [3, 8, 50]); improving the quality of data used to train and evaluate algorithms [41, 79] transforming training data into a form more amenable to learning [25, 60]; and increasing the size of training data [75].

---

[1] For the remainder of this paper, we refer to the combination of training and testing data, simply, as training data.

Although deep learning methods can perform outstandingly well with a minimal amount of preprocessing or explicit feature construction on unstructured data (e.g., image, audio, and text) [31], tabular data still pose a challenge to deep learning [12, 30] models [73], in which performance may strongly depend on the selected preprocessing strategy [31]. There have been some efforts to augment ML processes with domain knowledge at different stages (e.g. [10, 36]). These efforts include: creating handcrafted feature engineering to improve the accuracy of the models, using transfer learning, where patterns learned by one ML model are repurposed, fine-tuned, or distilled into smaller models; and using knowledge graphs with explicit relationships among entities to help provide domain knowledge [14, 44, 84]. Still, none of these works have attempted to enrich ML processes with rich domain rules (e.g., data cardinality, optionality of features or entities) as captured in organizational conceptual models.

Understanding patterns learned by ML models is often considered to be beyond reach, due to model complexity (black box). Representation learning targets methods for automatic discovery of human interpretable structures underlying data [9]. For instance, variational autoencoders as a generalization of principle component analysis (PCA) learn latent representations that capture the probabilistic structure of data [43]. The goal is to separate latent variables for better interpretation of the latent structure underlying data. Identifiability refers to methods that enable the determination of a unique, interpretable representation that can generate a dataset [70].

## 2.2. Process Transparency in Data Preparation

Machine learning tasks, such as identifying data sources, preparing data, and building and deploying models, are not systematized. However, several methods prescribe high-level activities for creating and deploying ML models. Two popular examples are the Cross-Industry Standard Process for Data Mining (CRISP-DM) and the Team Data Science Process (TDSP) Framework [29, 71]. Such methods require data cleaning as an important step in the ML process to improve accuracy or efficiency [22]. In existing methods, training data preparation is largely ad hoc [37] and based on intuition, judgment, or trial and error [25]. One survey of global ML practices concludes: "Everyone wants to do the model work, not the data work" [67] (p.6). The task of preparing the data before training ML models is "paradoxically…the most under-valued and de-glamorized aspect of AI" [67] (p.1). The result is a real risk that unconscious bias from the data scientists or in the underlying data will be embedded in ML models. This is a main motivation for the CMML method.

*Process transparency* explicitly articulates the data preparation techniques applied and describes how the tasks involved in a machine learning initiative are related. For example, data preparation tasks can involve transforming raw data into a usable form. These transformations include binning, normalization, imputation of missing values, feature engineering, and dimensionality reduction (e.g., [34]). There are accepted principles for working with datasets [74], as well as techniques for data cleaning [49] and feature engineering to generate useful (e.g., predictive) features from a raw dataset [25, 60, 85]. The techniques primarily focus on the statistical properties of the data; they do not provide an overall process for guiding data preparation to reflect domain knowledge. Manual feature engineering consists of extracting features from raw data and transforming them into formats suitable for the machine learning model. There is a general agreement among data scientists that the vast majority of time spent building a machine learning pipeline is allocated to feature engineering and data cleaning [85].

Most techniques used for feature engineering focus on ML performance without considering process transparency [60]. Process transparency can be further diminished when using Automated Machine Learning (AutoML) techniques, which automate many ML steps [49]. However, AutoML tools allow users to explicitly indicate their preference for either high performance (using computationally intensive and opaque transformations on training data) or improved transparency (using limited and more understandable transformations on training data) [35].

There is a well-recognized tradeoff between the objectives of process transparency and model performance [45]. For example, neural networks perform extremely well, but any representation learning contained within the hidden layer is comprised of weights and biases applied to the combinations of inputs at each node. As the number of hidden layers increases, subsequent layers contain nodes with weights and biases of the previous layers as inputs, making interpretability challenging.

In some cases, applying more complex and sophisticated transformations of the input data, whether created automatically by the ML pipeline (e.g., automatic feature engineering, representation learning) or by a data scientist (e.g., handcrafted feature engineering), may improve model performance. However, the lack of process transparency can decrease the effectiveness of such models. For example, while engaging in a feature engineering activity, data could be inappropriately manipulated by teams lacking domain experience, resulting in spurious or invalid relationships that can lead to inaccurate models. During the early part of the COVID-19 pandemic many tools unintentionally used data that contained chest scans of children who did not have a COVID diagnosis as their examples of what non-covid cases looked like. As a result, the AI learned to identify children, not COVID-19 cases. Similarly, researchers built models using data that contained a mix of scans taken when patients were lying down or standing up. Because patients scanned while lying down were more likely to be seriously ill, the AI learned wrongly to predict serious COVID risk from a person's position; not from whether they had developed severe pneumonia.[2] Data scientists must understand the domain well enough to discern among competing transformations to determine those that better represent the domain [60]. Process transparency can be improved by having data scientists select fewer and less opaque transformations [17].

### 2.3 Modeling

Conceptual modeling arose as a response to the need to understand and model the domain for which an information system or its components (e.g., database) is being developed and has evolved over time [5, 32]). Conceptual modeling describes "aspects of the physical and social world for understanding and communication" [58] (p.389). Entity-Relationship (ER) modeling is one of the most common approaches to creating conceptual models of data. The ER model, and its extensions, conceptualize a domain in terms of entity types that possess attributes[3] and participate in relationships with other entity types [21, 78]. Figure 1 illustrates the main constructs of the Extended ER (EER) model [78]: entity type, attribute, and relationship.

---

[2] https://www.technologyreview.com/2021/07/30/1030329/machine-learning-ai-failed-covid-hospital-diagnosis-pandemic/
[3] We refer to attributes within the context of conceptual modeling and as features within the context of machine learning.

An entity type represents a group of similar things of interest in an application domain and can be material or conceptual (Figure 1a). A relationship is an association between entity types (Figure 1c). Attributes are characteristics of entity types and relationships (Figure 1b). Composite attributes are comprised of other attributes (e.g., address is composed of street, city, state, and postal code). Derived attributes depend on the value of other attributes (e.g., age is derived from the date of birth and current date).

| Construct | Diagram Fragment Example |
|---|---|
| (a) Entity type: CUSTOMER | 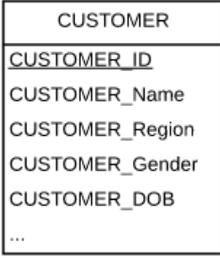 |
| (b) Attribute: CUSTOMER_ID CUSTOMER_Name CUSTOMER_Region CUSTOMER_Gender CUSTOMER_DOB | 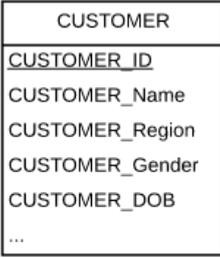 |
| (c) Relationship: One-to-one One-to-many Many-to-many | 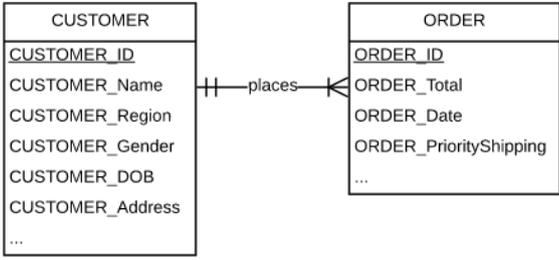 |
| | *Generalization/Specialization* 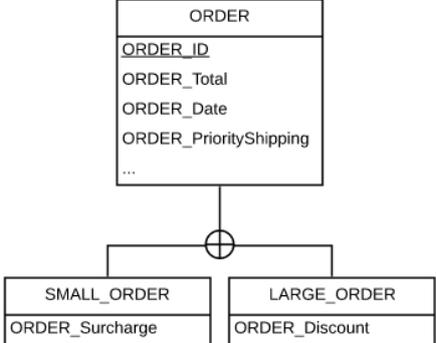 |

**Figure 1.** Extended Entity-Relationship (EER) Constructs Conceptual

Cardinalities capture the minimum number of entities (0 or 1) that can participate in a relationship and the maximum (1 or many). A zero cardinality indicates optional participation by an entity type in a relationship, while a minimum cardinality of one indicates mandatory participation. These constructs abstract knowledge of a variety of business rules or constraints that apply in a real-world application. Hierarchies of entity types are modeled by generalization/specialization. An entity subtype inherits all of the attributes of its supertypes and possesses one or more additional attributes or relationships.

When training data are extracted from an organizational database, the original data typically conform to some available conceptual model. However, in preparing data for training an ML model, information about entity types, relationships, and constraints is often lost when the data are reduced to a denormalized tabular structure.[4] Thus, conceptual modeling can affect ML model performance and facilitate transparency because using these entity types, relationships, and constraints can dictate the data transformation techniques for machine learning.

**2.4 Positioning within Conceptual Modeling and Machine Learning**

Research in information systems has long emphasized the importance of conceptually representing the real-world knowledge and logic needed for effective systems design [83]. These models provide a way to represent the facts, relationships, rules, and constraints of a domain. The ER community has contributed significantly to integrating conceptual modeling with machine learning. Nalchigar and Yu [59] proposed the use of conceptual models to support requirements elicitation. Lukyanenko et al. [51] propose conceptual models to support ML phases of the CRISP-DM cycle.[5] Maass et al. [53] and [52] propose a method for imposing the features weights of an ML model to enhance model interpretability. Maass and Storey [54] identified challenges associated with insufficient domain knowledge in ML, which conceptual models can potentially address. Corea et al. [24] repurpose DMN (Decision Model and Notation) decision tables as coalitional games and apply Shapley-value analysis to quantify each input's marginal contribution, providing a faithful, model-native explanation of complex decision logic.

---

[4] Machine learning requires data in a format similar to a flat file, which we call tabular data.
[5] CRISP-DM stands for Cross-Industry Standard Process for Data Mining and serves as a method that supports a structured and iterative approach to data mining projects.

Our work complements these theoretical foundations by providing specific operational guidelines for incorporating conceptual modeling principles into ML data preparation activities. In addition, research on Model Data Engineering (MDE) approaches complement these conceptual modeling efforts by systematically automating transformations and mappings between domain models and ML artifacts. This includes work by van de Reit 2008 [82], Burgueño et al. [16]; Bucchiarone et al. [15]; and Naveed [61]. To facilitate code development, "Model-driven software engineering and human–computer interaction design can help in abstracting machine learning technology" and enabling automated code generation" (Bucchiarone et al. [16], p.8).

## 3 CONCEPTUAL MODELING FOR MACHINE LEARNING METHOD

The Conceptual Modeling for Machine Learning (CMML) method, which we propose, incorporates domain knowledge contained in conceptual models into the data preparation process for machine learning applications. CMML is intended to improve machine learning model performance and process transparency for supervised machine learning tasks.

The foundation of the CMML method is based on the three main constructs of the Extended Entity-Relationship model: entity types, attributes, and relationships. CMML consists of an iterative process and five guidelines for using a conceptual model to prepare a dataset for training and testing machine learning models. An overview of CMML is shown in Figure 2. It requires two inputs: a conceptual model of a domain and an original raw dataset ($DS_0$) containing tabular data. After applying CMML, the outputs are one or more new training datasets ($TDS_n$), each used to train and test one or more ML algorithms.

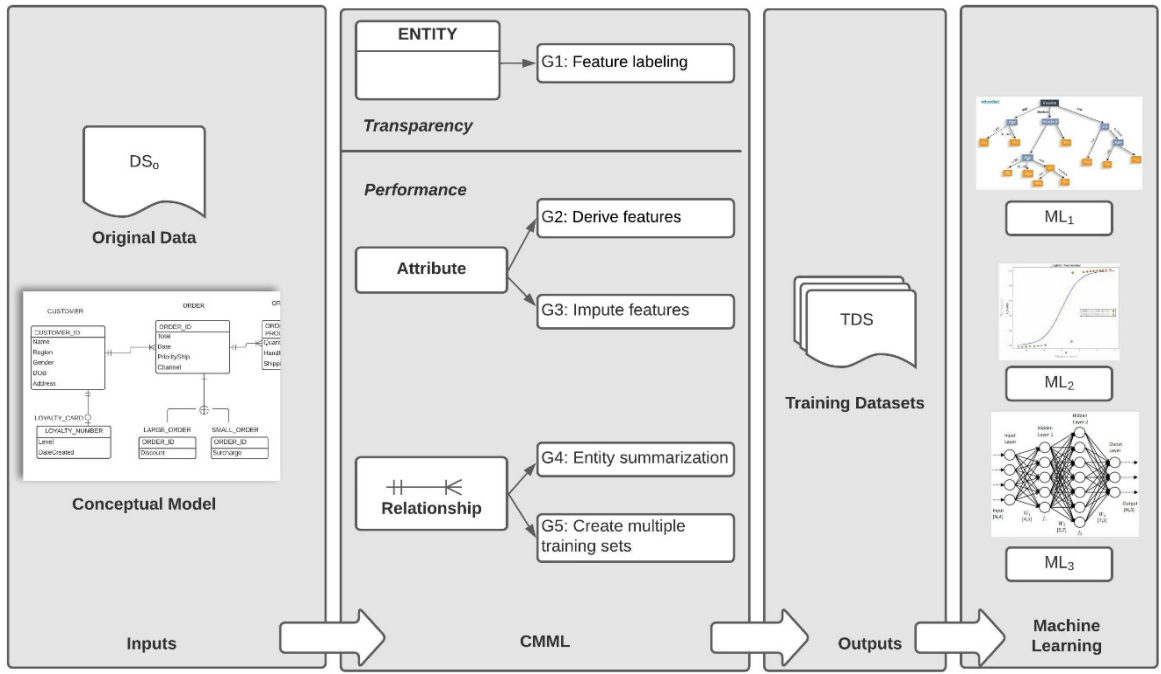

**Figure 2.** Conceptual Modeling for Machine Learning

## 3.1 Assumptions

CMML is based on two main assumptions. First, a conceptual model of the domain is available in the form of an extended entity-relationship (EER) diagram. Second, a dataset is available containing a target attribute and sufficient relevant features to construct a machine learning model. The following scenario, adapted from Khatri et al. [42], illustrates the development of the guidelines, with its corresponding EER diagram shown in Figure 3. The machine learning objective is to predict a customer's lifetime value (target attribute), which is derived from the total of all the orders for a customer. The guidelines are at the conceptual level, with specific examples at the physical level.

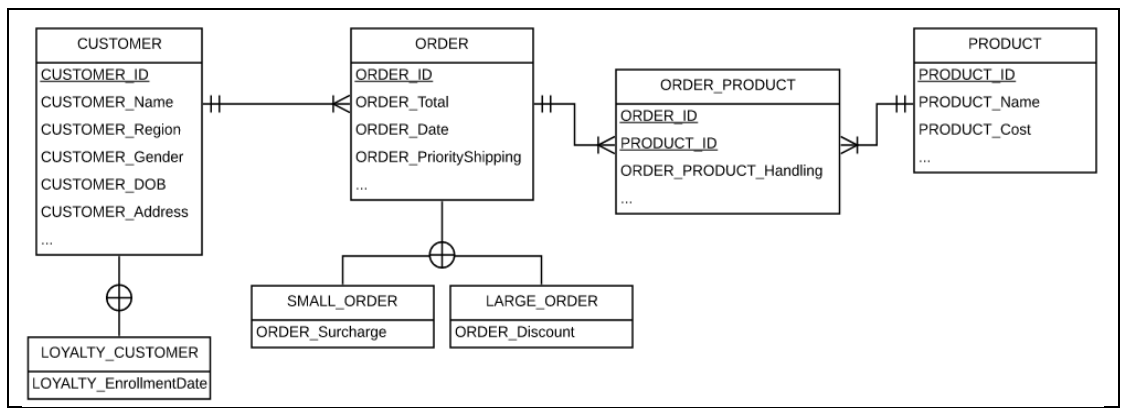

**Figure 3.** Conceptual Model for Customer Ordering (adapted from [42])[6]

The method iterates over the constructs of the EER model (entity type, attributes, and relationships) to preserve the domain knowledge expressed in the EER diagram during data preparation. We define important terms of our method in Table 1.

**Table 1.** Conceptual Modeling for Machine Learning Terms

| Term | Definition |
|---|---|
| Target attribute | Attribute of an entity type that can be used as a target feature in a training dataset for machine learning |
| Target-bearing entity | Entity type that contains the target attribute |
| Predictor entity | Entity type that contains attributes that will be used as feature variables in the ML training data |

### 3.2 Preparation

Entity types represent the classes of things of interest in a domain, such as CUSTOMER, VENDOR, and PRODUCT. Contemporary approaches to machine learning do not preserve knowledge about entity types or consider this domain knowledge when performing feature engineering [25]. This knowledge can be preserved by labeling the features. Although appending descriptive information to features is a standard ML practice, there is no systematic method for doing so. Appending the name of the entity type to the names of features can improve transparency by providing a consistent naming convention to indicate the lineage of the attributes. In Figure 3, the attribute Name appears in the PRODUCT and CUSTOMER entity types. Relabeling creates two differentiable features: PRODUCT_Name and CUSTOMER_Name.

Appending the name of the entity type is also helpful for maintaining lineage information of new features created via dimensionality reduction or feature engineering. Dimensionality, or feature reduction, is a feature transformation technique that reduces the number of input variables in training data. Dimensionality reduction combines several features into one using techniques such as clustering or principal component analysis. It is commonly used to deal with sparse data or to reduce multicollinearity among features [65].

---

[6] This model follows the crow's feet notation with many to many relationships represented by another associative entity, thus ORDER_PRODUCT 46.     Kroenke, D.M., et al., *Database Concepts*. 2010: Prentice Hall Upper Saddle River, NJ.

Labeling the new feature with the corresponding entities will improve process transparency. For example, in Figure 3, it is possible to cluster the demographic information for each customer to generate a new feature called "Demographic_Segmentation_CUSTOMER," which reduces dimensionality by removing more granular demographic information (e.g., address, region). The new feature name identifies the source entity type used in its derivation and includes it as a label. If more than one entity is involved in the transformation, the names of those entities are retained. For example, if we cluster customer and product information to generate a new feature, the name becomes: "Benefit_Sought_CUSTOMER_PRODUCT".

> ***Guideline 1 – Feature labeling.*** *All features maintain the named entities of origin. When conducting feature engineering, preserve the name of all corresponding entities.*

## 3.3 Attribute Guidelines

Whereas generally, attributes are turned into features in an ML pipeline, some valuable domain semantics can be lost in this process. We consider derived attributes and attributes with missing values to ensure we retain the semantics related to the attributes in an EER diagram. A derived attribute is an attribute whose value is not permanently stored in a database but calculated from the stored values of other attributes and/or additional available data (such as the current date) [39]. For example, CUSTOMER_Age can be calculated based on the difference between the current date and CUSTOMER_DOB.

As another example, we can compute the AVERAGE_ORDER_TOTAL for all the orders a CUSTOMER placed. In CMML, we apply the derived attribute concept to derive features from the available data. Since derived attributes do not exist in the database tables, but are important attributes of their respective entity, they must be calculated and added to the ML training dataset as new features. In CMML, data scientists must identify possible derived features and assess if they have the data to derive them.

> ***Guideline 2 - Derive features.*** *Identify derived attributes that can add more information than only using the attribute(s) in their raw form. For each derived*

> *attribute, create a new feature, appropriately and descriptively labeled, and compute the value of the new feature for each record in the dataset.*

A dataset that is used as input to ML can have missing values. A missing value that is not available or unknown but potentially knowable can be imputed using standard techniques, such as substitution, mean, interpolation, or extrapolation. However, a missing value might not apply to all members of an entity type. For example, CUSTOMER_level might not apply to all customers because some may not have joined a loyalty program. Missing values should be imputed *only* for those instances of the entity type for which the attribute is applicable. In this way, the entity-relationship model facilitates the identification of this case which might not be evident in the raw data.

> ***Guideline 3 - Impute features.*** *Impute missing values of a feature in a data set if the value is applicable but unknown.*

### 3.4 Relationship Guidelines

The CMML method exploits the relationships between entity types based on their cardinalities, which indicate whether the entity participation is *one to one (1:1)*, *one to many (1:N),* or *many to many (N:M)*. Machine learning algorithms are trained on a tabular flat-file dataset. Depending on the type of relationship, the data corresponding to each entity must be transformed to an adequate granularity to create a final dataset.

### 3.4.1 One-to-One (1:1) Relationships

In a one-to-one relationship, an instance of one entity type can be related to at most one instance of another entity type (and vice versa). These two tables (representing the entities) would be joined at the physical level. There is one record for each occurrence of an entity, so the data that corresponds to the instances of the relationship is already in a form that can be used by ML.

### 3.4.2 One-to-Many Relationships

In a one-to-many relationship, an instance of one entity type can be related to one or more instances of another entity type. As a result, there is a unit of analysis mismatch whereby the target-bearing entity is at a higher level of granularity than the predictor entities. To align the levels of unit of analysis, the data scientist will need to find a way to combine the data instances from each entity. Consider the customer/order scenario in Figure 4, where each customer has at least one order. The task is to predict

Customer LTV, a *derived* attribute for CUSTOMER calculated from information about the customer's ORDERs. Customer LTV appears on the one side of the relationship. The CUSTOMER entity type is a target-bearing entity because it contains the target attribute, CUSTOMER_LTV. It is also a predictor entity because it contains attributes that can be used as features (e.g., CUSTOMER_Age, CUSTOMER_Address, etc.).

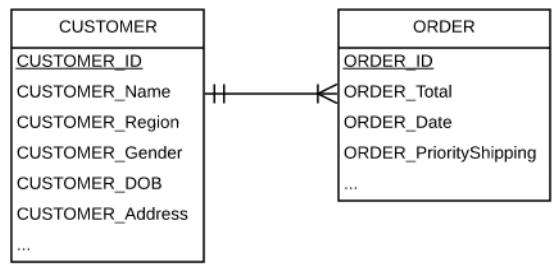

**Figure 4.** CUSTOMER-ORDER EER Segment

If we retain the granularity of this data at the order level, it will result in data from an entity type being repeated multiple times. The same customer information is duplicated for each order by a customer, including the CUSTOMER_LTV. It is not useful to have multiple rows of LTV for each CUSTOMER. Instead, these should be collapsed into one occurrence. Here, the target attribute is CUSTOMER_LTV and the features are anything from the CUSTOMER or ORDER tables. The repeating data is not included when there are several repeated rows for the target-bearing entity type (CUSTOMER).

Figure 5 illustrates data instances of $DS_0$ for the corresponding machine learning flat file based on the conceptual model in Figure 3. Customer 101 (CUST_ID =101) has placed four orders; thus, the customer attributes repeat (the order attributes are unique for each row). In contrast, Customer 400 placed one order. This file structure can over-represent the information from Customer 101 and lead to training case redundancy (Ohno-Machado et al., 1998).

| CUST_ID | CUST_Age (D) | CUST_Gender | ORDER_ID | ORDER_Total | ORDER_Date | ORDER_Channel | Order_PriorityShip | CUST_LTV (D) |
|---|---|---|---|---|---|---|---|---|
| 101 | 22 | M | 2778 | 100 | 11/1/16 | Phone | Y | 150 |
| 101 | 22 | M | 6591 | 50 | 1/17/18 | Online | N | 150 |
| 101 | 22 | M | 6391 | 17 | 1/15/18 | Online | Y | 150 |
| 101 | 22 | M | 8717 | 25 | 4/21/19 | Phone | Y | 150 |
| 400 | 27 | F | 2763 | 510 | 9/23/16 | Phone | N | 45 |
| 223 | 44 | F | 4574 | 32 | 12/10/18 | Online | N | 20 |
| 223 | 44 | F | 8185 | 87 | 4/7/19 | Online | Y | 20 |
| 398 | 87 | M | 9201 | 1025 | 5/3/19 | Phone | N | 35 |

**Figure 5.** Sample DS0 Data for Conceptual Model

When the target-bearing entity is on the one side of the relationship, we propose summarizing the records on the "many side" into a single row using an approach we call *entity summarization.* In Figure 5, information about ORDERS can be captured by creating new, summarized features (Duboue, 2020) that preserve information between instances of the target entity type (CUSTOMER) and associated instances of the predictor entity types (CUSTOMER and ORDER). For example, we capture the number of orders (summary feature) by creating a column called ORDER_Count. For Cust_ID 101, the ORDER_Count would be 1. Customer 101 has placed four orders, two online orders and two phone orders (two other summary features of nominal attributes), as shown in Figure 6.

| CUST_ID | CUST_Age (D) | CUST_Gender | ORDER_placed | ORDER_ChannelOnline | ORDER_ChannelPhone | CUST_LTV (D) |
|---|---|---|---|---|---|---|
| 101 | 22 | M | 4 | 2 | 2 | 150 |
| 400 | 27 | F | 1 | 0 | 1 | 45 |
| 223 | 44 | F | 2 | 2 | 0 | 20 |
| 398 | 87 | M | 1 | 0 | 1 | 35 |

**Figure 6.** Sample TDS$_1$ data for nominal attributes

For numerical attributes, we create summary statistics and other numerical transformations (counts, averages, sum, min, max, or ratios). For example, we conduct entity summarization on the total attribute of ORDER by using an average. Figure 7 demonstrates an excerpt of DS_1, which includes a new summarized feature, Order_Average. Note that, because of entity summarization, Customer_IDs 101 and 400 are now one row each.

| CUST_ID | CUST_Age (D) | CUST_Gender | ORDER_Average | CUST_LTV |
|---|---|---|---|---|
| 101 | 22 | M | 48.5 | 150 |
| 223 | 44 | F | 59.93 | 20 |
| 398 | 87 | M | 1025.5 | 35 |
| 400 | 27 | F | 510.7 | 45 |

**Figure 7.** Sample TDS1 Data for numerical attributes

> ***Guideline 4 - Summarize Entity.*** *When a one-to-many relationship creates data duplication, create a training dataset that removes training case duplication by summarizing the features on the many side of the relationship, creating counts for categorical variables and numeric summaries for continuous variables.*

### 3.4.3. Many-to-Many Relationships

Many-to-many relationships can have relationship attributes, whereas one-to-one and one-to-many do not. The resulting relationships are two one-to-many relationships and one associative entity that contains the keys of the entities involved in the original relationship. The attributes of the original relationship can exist but could have a null value. For example, ORDER_PRODUCT links ORDER and PRODUCT entities and has relationship attributes: quantity, handling, and shipping. No further action is necessary since the many-to-many relationship has been split into two one-to-many relationships, so the guidelines for one-to-many can now be followed.

### 3.4.4. Generalization/Specialization

Generalization/specialization is a special entity type structure that captures the semantics of classes and subclasses that occur commonly in the real world. CMML exploits the generalization/specialization structure to identify potential quality issues in (subtypes) cases where the cardinality of the data is not one-to-one. In Figure 3, the $\oplus$ symbol below the ORDER entity type indicates a generalization of the disjoint subclasses, SMALL_ORDER and LARGE_ORDER, with the former having a surcharge attribute and the latter a discount attribute.

Without this domain knowledge, a data scientist might have a dataset that combines SMALL_ORDER and LARGE_ORDER and might impute values for missing data without regard to the subtype. In this example, a value for surcharge may not be applicable because the ORDER with the missing value is a LARGE_ORDER, which makes any data imputation (e.g., using an average) inappropriate. If the missing values are not imputed, this reduces the training sample size for machine learning because some algorithms (e.g., neural networks) cannot handle missing values. However, if missing values are imputed, this creates noise.

> **Guideline 5 - Create multiple training datasets.** For *generalization and specialization, build multiple training datasets (TDS$_n$), one for each subtype, to train distinct models.*

When using CMML in the presence of specialization, the subtypes are not always disjoint. Subtype overlap is where there are instances that are members of more than one subtype. These instances should be duplicated such that they are present in each TDS$_n$ of which they are a member. For example, while most products might only be of type consumer, retail, or manufacturer, a particular product could be a member of all three of these subtypes. Although each subtype captures only attributes specific to it, these overlap instances will have values across multiple subtypes. In all cases where there are subtypes, build each subtype's TDS$_n$ using: (1) only the instances that are members of that subtype, and (2) with only the values specific to that subtype. The output for the discussed example would be three $TDS_n$: ($TDS_1$) consumer product; ($TDS_2$) retail product; and ($TDS_3$) manufacturer products. A product that is a member of all three of these would be present in each, but with only those attributes that apply to each respective subtype. These data can be used to create a separate model for each subtype.

**3.5 Application of guidelines**

Machine learning tasks and the data available for these tasks can vary. It is unlikely that a given machine learning task will require the use of all guidelines (e.g., there could be no missing values, or derived attributes). Therefore, combinations of guidelines may be used as needed. However, all tasks should adhere to Guideline 1, which seeks to maintain transparency by including the name of the entities of origin. Data scientists have multiple ways to receive data. These guidelines can be directly applied if they obtain the data from a relational database. However, if the data comes as a flat file (or files), such as when getting data from a data lake which commonly occurs, then the benefits to ML occur as summarized in Table 2.

**Table 2.** Summary of Guidelines

| Guideline | Benefits for ML |
|---|---|
| *G1.* **Feature labeling**. *All features maintain the named entities of origin. When conducting feature engineering, preserve the name of all the corresponding entities.* | Ensures consistent naming convention and greater traceability of features. |
| *G2.* **Derive features**. *Identify derived features that can add more information than only using the variable(s) in their raw form. For each derived attribute, create a new feature, appropriately and descriptively labeled, and compute the value of the new feature for each record in the dataset.* | Engineering features that represent the underlying problem domain. |
| *G3.* **Impute features.** *Impute missing values of an attribute in a data set only if the value is applicable but unknown.* | Domain-aware and semantically appropriate imputation of missing values. |
| *G4.* **Summarize entity**. *When a one-to-many relationship creates data duplication, create a training dataset that removes training case duplication by summarizing the features on the many side of the relationship, creating counts for categorical variables and numeric summaries for continuous variables.* | Removal of repetition in some of the attributes of the data.<br><br>Merging datasets that are at different granularities. |
| *G5.* **Create multiple training datasets**. *For generalization and specialization, build multiple training datasets (TDSn), one for each subtype, which will be used to train distinct models.* | Ensuring imputation of missing features that only apply to certain entities. |

## 4 Evaluation

We demonstrate and evaluate the CMML method by applying it in a real-world context to show how it can improve ML performance and process transparency, and eliciting an assessment of the method from data scientists.

### 4.1 Application to Real-world Context: Foster Care

To assess the method within a real-world context, we obtained data from two foster care organizations in the United States, both motivated by an interest in using ML to improve their operations.[7] One author has deep domain knowledge due to extensive work with several foster care agencies for 15 years. Figure 8 shows a conceptual model that represents a typical foster care domain.

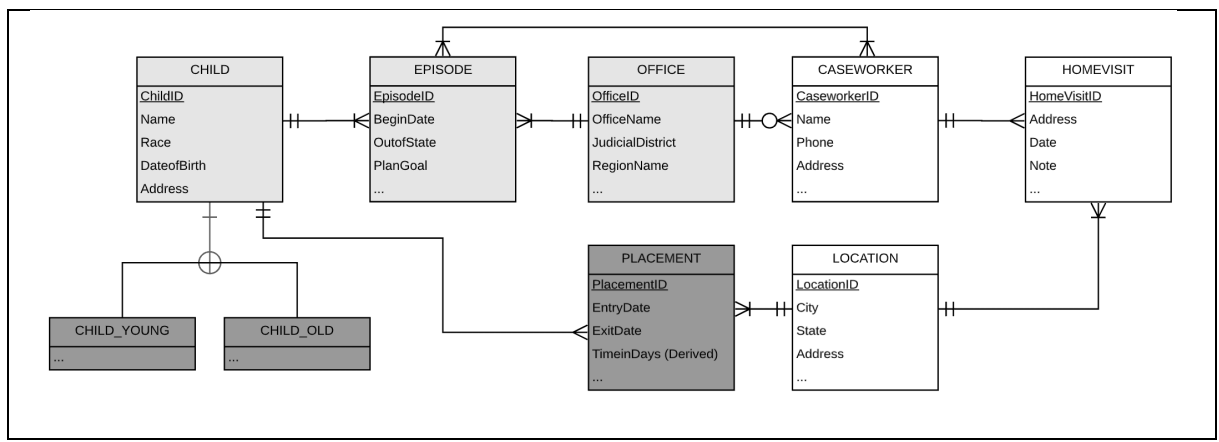

**Figure 8.** Application: Conceptual Model of Foster Care Case Management

For the foster care organizations we worked with, a child's overall length of stay is an *episode*. Episodes are managed within a foster care office by assigning them to caseworkers. A child could have more than one episode reported in the foster care reporting system. For example, a child could be permanently placed (episode 1) but later return to foster care (episode 2). Caseworkers must conduct a home visit (the location where a child is placed) periodically (which could vary depending upon the jurisdiction). A location can have more than one foster child placement.

A significant challenge in foster care is identifying the characteristics of a foster care child or case (e.g., age, number of siblings, placement type) that predict the length of stay [7]. A second challenge is detecting the over-prescription of psychotropic medications, medications prescribed to help children cope with behavioral problems, such as attention-deficit/hyperactivity disorder, depression, bipolar disorder, and psychotic disorders. Failure to identify at-risk children is highly problematic because adverse outcomes can include severe consequences, including death.

---

[7] A foster care system in the United States is a temporary arrangement in which adults provide care for a child whose birthparents are unable to care for them. Foster care is usually arranged through the courts or a social service agency. The goal for a child in the foster care system is to reach a permanent living arrangement. In foster care, a child lives with a relative or non-relative adult who has been approved by the State, or by an agency licensed by the State.

Two datasets were used to evaluate CMML. The first (case 1) was a structured database describing the placements of children in the foster care system. These data were used to build machine learning models to predict a child's length of stay (episode) in the system. The portion of Figure 8 that appears in dark gray is relevant to this use case. The second (case 2) was a database containing unstructured text notes about caseworkers' visits to foster care homes, which was used to build models to predict whether a child is prescribed psychotropic medication. The portion of Figure 8 that appears in white is relevant to this use case.

In case 1, the machine learning task was to evaluate the performance of models that predict the length of an episode. In case 2 (unstructured data), the machine learning task was to evaluate the performance of models that classify children as taking psychotropic medication(s) versus those who are not.

**4.1.1 Case 1 – Prediction of Length of Stay (Estimation)**

The data used to predict a foster care child's length of stay (episode) was extracted from a secured data portal that collects data on children entering and leaving the foster care system in real time. The portion of Figure 8 in dark gray and light gray illustrates the relevant part of the domain for applying the CMML method[8]. The data was stored in a relational database and exported into a flat file. The resulting dataset contained 25,462 placement records, representing 4,437 episodes and 9,942 children from January 1, 2015, to March 31, 2019. The target attribute was the length of an episode.

An example of $DS_0$ is given in Figure 9. The objective is to build a model that estimates the length of stay for a foster child. For example, foster care children reunified with parents or primary caretakers have a median length of stay of 66 days. In contrast, children placed in care via a court ordered placement have a median length of stay of 238 days.

---

[8] Not all attributes are pictured in this diagram.

| Case Plan Goal | Placement Type | ... | PlacementOutcome | Entry_Age | LOS-Target |
|---|---|---|---|---|---|
| Adoption | Foster Family Home (Non-Relative) | ... | Emancipation | 16 | 649 |
| Planned Permanency Living Arrangements | Group home | ... | Reunif With Parents Or Prim Caretakers | 8 | 55 |
| Reunify with Parent(s) or Principal caretaker | Foster Family Home (Non-Relative) | ... | Unknown | 10 | 45 |
| ... | ... | ... | ... | ... | ... |
| Guardianship | Runaway | ... | Court Ordered Placement | 17 | 678 |
| Live with Other Relative(s) | Protective Supervision | ... | Discharge/No Further Agncy Spvrsn/Jusris | 8 | 140 |

**Figure 9.** Snippet of dataset $DS_0$

After creating $DS_0$, we applied the CMML guidelines to transform it and create several training data set ($TDS_n$) versions, as shown in Table 3. Each version results from applying one or more CMML guidelines to $DS_0$ (in addition to guideline 1).

**Table 3.** Datasets Description, Case 1 (predicting length of stay)

| Dataset Summary | Guidelines Applied |
|---|---|
| *Dataset*: **$DS_0$**<br>*Row Representation*: **One placement, the lowest granularity available for the task.**<br>*Sample Size*: **25,462**<br>*Target attribute Mean*: **704.78**  *SD*: **414.89**<br><br>*Guidelines Applied*: **N/A** | N/A – the original raw dataset (DS0) contains the data in tabular format for Case 1: predicting length of stay. |
| *Dataset*: **$TDS_1$**<br>*Row Representation*: **One case which is comprised of one or more episodes.**<br>*Sample Size*: **4,437**<br>*Target attribute Mean*: **440.47**  *SD*: **340.70**<br><br>*Guidelines Applied*: **Guidelines 2 and 4** | **Guideline 2: Derive features**– When summarizing the data (per guideline 4), several new derived attributes were created, such as counts of different placement types, number of total placements, one-hot encoding to capture all the different types of placements (such as group home, relative, foster parent).<br>**Guideline 4: Summarize entities** – The target-bearing entity was EPISODE. The data in $DS_0$ was aggregated for its respective case. Note that children could have more than one episode. We kept data aggregated to the episode level. Thus, a child having more than one episode could appear multiple times. |
| *Datasets*: **$TDS_2$** –Younger, **$TDS_3$**-Older<br>*Row Representation*: **Subsets of dataset consistent with the age of the child upon start of the episode**<br>*Sample Size $TDS_2$-Younger*:**12,436**<br>*Target attribute Mean*: **645.39**  *SD*: **381.51**<br>*Sample Size $TDS_3$-Older*:**13,026** | **Guideline 5: Create multiple training datasets.** We separated the data into two datasets based on standard, well-accepted child subclasses (age 7 or younger, 8 and older) (Courtney et al., 1996; Kadushin & Martin, 1988). |
| *Target attribute Mean:* **761.48**     *SD:* **436.98** <br><br> *Guideline Applied*: **Guideline 5** | |

To demonstrate the robustness of our method, we used both popular ML algorithms (typically used by experienced ML teams) and a commercial AutoML tool [19]. For the former, we selected Deep Learning algorithms [30], Random Forests [11], Gradient Boosting Machines (GBM) [28], and Light GBMs [40]. These are well-known and generally effective ML techniques for tabular data [29]. We used Python, and open source libraries sci-kit learn and h2o.ai [62]. For AutoML, we created models using a popular commercial tool, called H2O Driverless AI [17]. This tool automates the process of algorithm selection, feature engineering, and hyperparameter tuning.

Building actionable machine learning models involves comparing the performance of different models while considering different algorithms and hyperparameter tuning [81]. The modeling was carried out using a general-purpose GPU compute Amazon AWS cloud instance with an Intel Xeon E5-2686 v4 (Broadwell) processor, 61 Gb RAM, NVIDIA K80 GPU (12 Gb), and h2oai-driverless-ai-1.8.5 AMI (Amazon Linux). We used RMSE (Root Mean Square Error) to measure predictive accuracy based on out-of-sample assessment (cross-validation) [72].

The RMSE and $r^2$ are standard ways to assess an ML model's estimation accuracy. Table 3 shows these results. Bolded items indicate a significant reduction in RMSE or an increase in $r^2$ compared to the $DS_0$-based model. We also calculated the normalized RMSE (nRMSE) to ensure an appropriate comparison of RMSE across different datasets. The range of all the datasets was identical (min=0, max=1550).

As seen in Table 4, the results show that models built using $TDS_1$, which applies Guideline 2 (derive features), and Guideline 4 (entity summarization) to $DS_0$, consistently outperformed models built using $DS_0$, with an average increase of 23.8% in explained variance across the various models and an average reduction in RMSE of 24 (7.4%). The application of Guideline 5 resulted in two subsets of the data: $TDS_2$ (younger children) and $TDS_3$ (older children). There was an improved performance for the $TDS_2$ across all models, and improved performance for the $TDS_3$ AutoML model. Combining the results for TDS$_2$ and TDS$_3$ for each model showed a 19% increase in explained variance with the AutoML model, however equivalent performance across the other four models to DS$_0$.

**Table 4.** Model Comparisons ($r^2$ and RMSE)

|  | **Deep Learning** <br> *RMSE NRMSE ($r^2$)* | **Random Forest** <br> *RMSE NRMSE ($r^2$)* | **Gradient Boosting Machine (GBM)** <br> *RMSE NRMSE ($r^2$)* | **Light GBM** <br> *RMSE NRMSE ($r^2$)* | **AutoML** <br> *RMSE NRMSE ($r^2$)* |
|---|---|---|---|---|---|
| DS$_0$ | 356.31 .23 (0.26) | 307.45 .20 (0.45) | 316.57 .20 (0.42) | 320.38 .21 (0.40) | 244.44 .15 (0.47) |
| TDS$_1$ – guidelines 2 and 4 | **218.06 .14 (0.59)** | **196.87 .13 (0.67)** | **208.44 .13 (0.63)** | **208.22 .13 (0.63)** | **85.96 .06 (0.67)** |
| TDS$_2$ –Younger Guideline 7 | **322.06 .21 (0.29)** | **290.52 .19 (0.42)** | **286.08 .18 (0.44)** | 305.45 .20 (0.36) | **177.83 .11 (0.59)** |
| TDS$_3$-Older Guideline 7 | 393.75 .25 (0.19) | 342.65 .22 (0.39) | 342.31 .22 (0.39) | 348.50 .22 (0.36) | **101.18 .07 (0.73)** |

Although the RMSE and $r^2$ values suggest practical value, the paired prediction absolute errors were compared statistically between the best performing non-AutoML model for DS$_0$, Random Forest, and the various TDS$_n$ datasets using Wilcoxon Signed-Rank Test. See Table 5. DS$_0$ was also paired with the combined results of the TDS$_2$-Younger and TDS$_3$-Older, because these are disjoint subsets of the data contained in DS$_0$. The p-value presented is two-tailed. The null hypotheses that the TDS$_n$ versions produced from DS0 using CMML had no effect on prediction errors (compared to DS0) were all rejected at $\alpha$ = .05. Based on the evaluation, we conclude that applying the CMML method can improve the performance of ML models.

**Table 5.** Results of Wilcoxon Signed-Rank Test

| **Paired Comparison** | **Sample Size (n)** | **SD ($\sigma_{T^+}$)** | **z-score** | **p-value** |
|---|---|---|---|---|
| DS$_0$ vs TDS$_2$-Young + TDS$_3$-Older | 25462 | 1172900 | -15.85 | <0.01 |
| DS$_0$ vs TDS$_1$ | 12438 | 400462.1 | -89.61 | <0.01 |

To demonstrate transparency improvements following CMML, we applied Guideline 1 globally to label the features with the relevant entity names (e.g., CHILD, EPISODE, PLACEMENT). In Figure 10, we show the final features (original or engineered) used in the model with their respective entity names (after applying Guideline 1). The top five final features used in Case 1 are shown in Figure 10 in descending order by importance. If no feature engineering was performed, the feature has "Original" identified in its label. Note that even for the transformed features, we now specify the name of the entity type Placement (e.g., PLACEMENT_OutcomeTrialHomeVisit), following Guideline 1. As readily seen from this illustration, PLACEMENT is the key predictor entity, something that would not be obvious by considering the features in isolation of the entities. Later, we demonstrate beliefs about transparency improvements based on CMML by conducting a focus group with target practitioners.

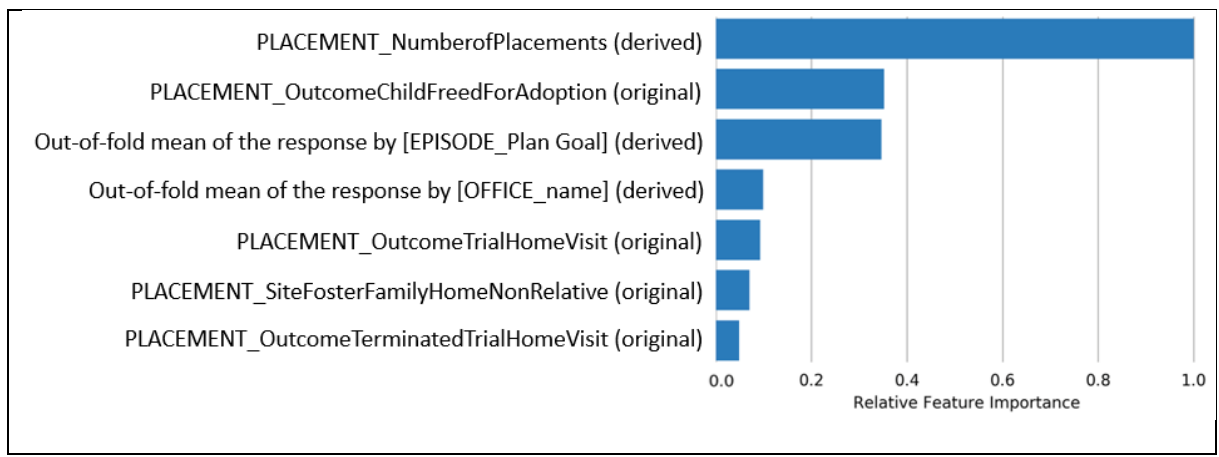

**Figure 10.** Feature importance after applying Guideline 1 – feature labeling

### 4.1.2 Case 2 – Prediction of Psychotropic Prescription (Classification)

This second machine learning project aimed to help caseworkers identify children taking psychotropic medication. The organization previously used a random selection method to identify potential positive cases of psychotropic drug use. The organization faced new mandates requiring monitoring of children taking psychotropic medication that could be overprescribed. This additional mandate added a new task to caseworkers' workload: analyzing random samples of existing home-visit notes, which document the interaction between the caseworker and the child at their home, to identify children who might be taking these medications. The organization contacted the researchers to seek assistance in identifying cases where children were receiving these medications.

The portion of the EER diagram relevant to this use case is highlighted in Figure 8 in white and light gray. The researchers collected a data sample from 852 children. The domain experts (caseworkers) manually verified each child as being on psychotropic medication or not. Next, the researchers collected the case note texts (from the caseworkers) for this sample over six months. Out of the subset of 852 children, 92 cases (10.8%) were of children taking psychotropic medication. The original and transformed datasets are described in Table 6. As the EER shows, the home-visit note reported by the caseworker is at the home level, and a home might have more than one child. Trying to detect evidence of a child taking psychotropic medication is challenging when there is information in a note about several children, some of whom may not be using psychotropic medications. Another challenge is that there are several case notes per home. We applied Guideline 4 by combining all the notes for one child and Guideline 5 by creating separate datasets, including one where there is a one-to-one mapping of a child to their home-visit note (with annotations that are specific to that child, e.g., a child is taking 5 mg of Adderall twice daily, 3-4 hours apart).

**Table 6.** Dataset Description – Case 2

| Dataset Summary | Guidelines Applied |
| --- | --- |
| *Dataset*: $DS_0$<br>Row Representation: **One home visit note, the lowest grain available for our task**<br>*Sample Size*: **1,545**<br>*Guidelines Applied*: **N/A** | N/A as this was the original data – the home visit notes from caseworkers for a sample of children verified by them to be taking psychotropic medication, or not taking psychotropic medication, during a period of 6 months (December $1^{st}$ – May $30^{th}$). |
| *Dataset*: $TDS_1$<br>*Row Representation*:<br>**All home visit notes about one child**<br>*Sample Size*: **852**<br>*Guidelines Applied*: **Guidelines 4 and 5** | **Guideline 4: Summarize Entities** –The target attribute is at the CHILD level, so the individual home visit notes were aggregated to form one corpus per child. There were multiple cardinalities present in the data. Many of the home visit notes were about one child, while some were about several children.<br>**Guideline 5: Create multiple training datasets**—These were split into two datasets: ($TDS_1$-Single) homes with one child; and ($TDS_2$-Multiple) homes with more than one child. Due to the nature of the research question (a child taking prescribed psychotropic medication), we dropped $TDS_2$ as these notes were not about a child, but rather many living in one household. |

We applied traditional ML techniques using Python and open source libraries of sci-kit learn and h2o.ai [63] to convert text into a format amenable for machine learning. Natural language processing (NLP) techniques (e.g., tf-idf, bi-grams, word embeddings) [26, 56] were applied to the initial unstructured data set to derive $DS_0$ (structured dataset). $DS_0$ was used to extract features prior to applying the CMML method and building the models. $TDS_1$ was processed using the same operations as $DS_0$. The out-of-sample method uses a 5-fold cross-validation. We used common evaluation metrics of recall, precision, and F-measure to evaluate the classification performance results. Table 7 shows the results using a popular and efficient machine learning algorithm, Light GBM [40]. We also applied other machine learning algorithms, such as a random forest and a neural network to the transformed dataset and obtained similar results.

**Table 7.** Unstructured data model comparison (F1-measure)

| Data | Precision | Recall | F-measure |
|---|---|---|---|
| $DS_0$ | 40.56 | 59.28 | 48.17 |
| $TDS_1$-Single | 38.89 | 84 | 53.17 |
| Z-score (one-tailed) | z = 0.2313 | z = 2.411** | |

*\*\* indicates significance level of p-value < .01*

Although there was a slight decrease in the precision of the ML models built on the transformed data set, the difference in precision was not statistically significant (p=0.41). The recall metric significantly improved (p = 0.008) with the transformed data set, and the F-measure difference was 5%, indicating that the application of our guideline improved performance, and the improvement was statistically significant [2]. We are most interested in the recall metric due to the importance of identifying as many positive cases of psychotropic drug use as possible for further evaluation by the organization. Although this emphasis on true positives (identifying more of the children taking psychotropic medication) is desirable in our case, it can also increase the rate of false positives (children wrongfully identified as taking psychotropic medication). This potentially undesirable increase is captured in the precision metric, which was not significantly decreased to achieve our desired results.

### 4.2 Assessment by Data Scientists

To assess the applicability of CMML, we engaged data scientists who would potentially benefit from using the CMML method using focus groups. Focus groups "can be very informative and lead to better and more relevant management implications" because they provide direct interaction with participants [80] (p.4). Focus groups allowed us to obtain feedback that might not have surfaced with other evaluation strategies, such as one-on-one interviews, surveys, or lab experiments [47] and have been used successfully in prior information systems research (e.g., [18, 68]).

We followed an established approach to focus group design and recruitment [80]. The goals of the focus group were to: (1) introduce participants to applying conceptual modeling in preparing a dataset for machine learning by applying the CMML method; and (2) evaluate the applicability and potential usefulness of the five guidelines and discuss how CMML can help systematize data preparation for machine learning. We engaged 15 professional data scientists who are potential beneficiaries of applying the CMML method in their organizations. The average age of participants was 34 years old with five years of experience on data-related projects. Participants were divided into three groups to facilitate discussion.

Each participant received two documents, the first describing the guidelines with the running example adapted from Khatri et al. [42] and the second containing a summary of the guidelines. Appendix A provides an overview of the focus group protocol. Each focus group was recorded and transcribed. The coding was completed in two rounds. Two coders systematically reviewed one-third of the focus groups' transcripts to identify sections relevant to evaluating the potential usefulness of the five guidelines. A Pooled Cohen's Kappa [23] inter-rater agreement of 0.66 was achieved in the first round, a good agreement between the coders [57]. After resolving the initial disagreements and completing the coding of the transcripts, an inter-rater agreement of 0.81 was achieved, indicating excellent agreement between coders [57].

Overall, our data analysis offered evidence for the utility of the CMML method. Early in the focus group sessions, participants remarked that although machine learning is a structured process with standard machine learning methods (e.g., Knowledge Discovery in Databases (KDD), CRISP-DM, Sample/Explore/Modify/Model/Assess (SEMMA)), the activities performed within each step are often ad-hoc. One participant stated that understanding the domain using a conceptual model can help create a "checklist that I can sort through as I am preparing the data…" since "it makes these steps more structured, for example, how to aggregate data or to consider null values, which all happens when you are dealing with real-world data." Another participant added "dealing with the domain [data] right upfront will make it easier down the road because you know and understand what you are working with" and "having the conceptual model helps you understand the data better. This contextual knowledge helps avoid misrepresenting the domain."

In general, the better the representation of a domain, the better the performance and transparency of the models. One participant stated, "by getting the structure right and properly reflecting the most robust version of the real-world structure, you get the best working model… avoiding making assumptions of the data prevents bigger issues in the outcome." In highly regulated fields in which complex models can affect lives, having a transparent process is essential. A manager acknowledged this by stating that "a regulator or auditor can ask for the data lineage, so traceability is important and making sure the data was used correctly and the validation they are getting from their models are consistent," providing further evidence that the CMML method improves transparency.

Participants appreciated how applying Guideline 1 can effectively interpret the model outputs: "as you would not want to accidentally mismatch attributes of one entity to that of another entity." Another participant, referring to Guideline 4, emphasized how common it is to miss relationships between entities in a raw dataset, particularly in a target-bearing entity type. He stated, "the target entity type, whether it is on the one side [and] exploring ways to aggregate the *many side*, is a key preparatory step in modeling efforts." Participants also stated the value of understanding entity type specialization in discussions about Guideline 5: "I think it is an interesting point about the difference between a missing value and an optional value; two different things referring to Guideline 5. You know, if it's blank, I don't know today, but if it is [truly optional], if it's been documented, then imputation is risky. You should be thoughtful and decide if it is appropriate before you impute."

Participants identified a few challenges in applying the guidelines. One participant stated, "I feel like I would struggle to find specific instances of things that could cause bias without actually looking straight at the data instead of relationships." Another participant indicated, "there are challenges when you are introducing a competing framework. Some people have already adopted practices [and] if you have never touched SQL, what is an optional attribute and how do you think about it in terms of moving forward with that?" Another participant stated that following a conceptual modeling approach to machine learning in some cases "may make your models performance lower because you are being more honest."

Overall, the focus groups offer compelling evidence of the potential usefulness of the CMML method. Many of the data scientist participants planned to adopt these guidelines early in the machine learning process, with one participant stating that doing so "would clearly save efforts later." The participants confirmed that using the guidelines could result in improved transparency and improved performance.

## 5 DISCUSSION AND IMPLICATIONS

The main contribution of this research is showing how conceptual models, which are traditionally seen as tools to support database design and information systems development, can be used in preparing data for machine learning to capture and use domain knowledge. Our approach can increase machine learning model performance and process transparency, and has implications for ML theory and practice, conceptual modeling, and information systems research.

The CMML method provides a mechanism to incorporate domain knowledge from conceptual models into the machine learning process. The empirical analysis demonstrates that incorporating conceptual models in the data preparation process can improve machine learning performance. Improvements were shown in both structured and unstructured data in two different real-world applications. Since both cases were aimed at improving case management in a foster care system, the improvement observed in machine learning performance has been shown to have a concrete, real-world impact. Finding ways to improve ML results with limited data has created a push for "small data" research over the past decade [19]. Although a high level of performance can be attained from an abundance of data, our work demonstrates how conceptual modeling can be effective in improving performance without necessitating more data.

The application of CMML has further been shown to improve the process transparency of machine learning, a relatively under-studied issue but one that is gaining steady interest [54]. As machine learning becomes more widespread, so are concerns that the process of building models remains opaque and without sound methodological grounding [37, 38]. Our results show consistent and robust evidence for the benefits of using CMML to improve process transparency.

This finding is significant as CMML improves performance, thereby showing that it is unnecessary to reduce performance to increase transparency, at least in some settings. They provide a compelling argument for introducing conceptual modeling as an integral component of machine learning projects and in machine learning curriculum. The latter is to better support data scientists in implementing the method, which was a need identified through our focus groups and analogous to the need to understand how to apply conceptual modeling for database design.

Because our research is applied in the real-world foster care setting, it has organizational implications. Today, many ML projects proceed without any conceptual modeling support. Machine learning projects should be adjusted and include conceptual modeling as a tool to support them. Embedding domain knowledge is vital because machine learning projects are increasingly used in societally critical domains, such as foster care, where decisions can dramatically affect people's lives. Thus, its deployment should be carried out in a more transparent, repeatable, and systematic manner.

Our work is also significant for the broader adoption of ML by contributing critically missing elements in automated machine learning (AutoML). The high degree of automation in AutoML allows non-experts to use machine learning models, thereby making ML accessible to many more people [49]. However, with increasing accessibility to ML, new problems emerge as non-expert users rely on algorithmic procedures. They might have little understanding or control needed to build models [27]. Currently, AutoML tools view performance and transparency as tradeoffs.

Furthermore, AutoML tools lack knowledge of specific business domains, a limitation conceptual models are uniquely positioned to address [76]. It should be possible to incorporate methods such as CMML into the pre-processing routines of AutoML tools and machine learning interpretability frameworks [20, 66]. Doing so would then permit non-technical users to upload a conceptual model with the corresponding dataset to communicate valuable domain knowledge and constraints to the automated tool. Conceptual models, which are designed to be accessible to non-technical people and serve as a communication tool [48, 58], could, in principle, become a means of better managing data preparation tasks in AutoML.

Although we demonstrated several benefits of our method, there can be cases where adding domain knowledge expressed in a conceptual model leads to degraded ML model performance. For example, this could happen if the conceptual model is outdated or incomplete and, therefore, inconsistent with the organization or domain it is intended to represent. A conceptual model represents articulated and agreed-on domain knowledge. However, as business processes are carried out and data collected, there might be deviations from the rules expressed in a conceptual model. The real-world data might not conform to the conceptual model; nor might the model be the best representation of domain knowledge. In addition, the value of conceptual models might be limited to situations where there is insufficient data to extract all relevant domain knowledge. An additional challenge is to convince practitioners to change their current practices to apply the CMML method.

A significant assumption in our approach is that there exists an agreed-upon conceptual model for the data describing the domain of interest. This might not be the case, especially if the data come from unstructured sources or include user-generated content. Furthermore, if the available dataset was developed by integrating data from multiple sources, it might reflect inconsistent domain knowledge that must be reconciled before the model can be used effectively. These issues motivate more research in conceptual modeling on view reconciliation and automatic extraction of conceptual models from organizational documentation.

The CMML method can extend beyond structured data to support diverse data types commonly integrated within modern data fabric architectures. As demonstrated in Case 2 (Section 4.1.2), CMML can accommodate unstructured textual data and can be applied to other unstructured formats such as images, audio recordings, or video. In data fabric environments, which serve as semantic integration layers across heterogeneous data sources, conceptual models provide contextual domain knowledge. Entity types identified in the conceptual model provide context for feature extraction from unstructured sources, whether through natural language processing, computer vision algorithms, or audio analysis techniques. The integration of data fabric technologies that automate metadata extraction and knowledge graph construction across heterogeneous sources can further enhance the CMML method, maintaining conceptual model alignment throughout the ML pipeline while preserving domain semantics. This integration is particularly valuable as organizations adopt AI-powered data fabric solutions that automatically discover relationships and metadata across heterogeneous sources.

The CMML method complements these technologies by providing methodological guidance for preserving domain semantics throughout increasingly automated data preparation workflows, addressing a critical challenge in contemporary machine learning applications. CMML does not replace existing pipeline tooling. Instead, it grounds data fabric tools in an explicit conceptual schema, ensuring that automated ingestion, transformation, and governance tasks remain transparent and aligned with the domain.

The substantial promise of infusing conceptual modeling knowledge into ML motivates further investigations into the boundary conditions of this approach and ways to overcome some of its potential limitations for special applications and scenarios. Our intent was to demonstrate that CMML can improve ML process transparency and performance. However, we did not evaluate the entire method. Indeed, since not all guidelines apply to a given situation, optimal guideline selection must be specific to each context and the corresponding conceptual model. From our assessments, we conclude that it is possible to improve ML models' process transparency and performance, but more work is required to consider how to prioritize the guidelines for specific use cases and evaluate CMML as a whole.

## 5.1 Expansion of Conceptual Modeling

Using conceptual modeling to improve performance and transparency in machine learning expands conceptual modeling research in information systems. The CMML method is one way to apply conceptual modeling to ML. Further research is needed to apply other conceptual modeling knowledge to different aspects of ML practice. For example, recent advances in generative AI (GenAI) technologies have significantly enhanced metadata extraction capabilities relevant to conceptual modeling. Large language models (LLMs) can now automatically identify entities, relationships, and attributes from unstructured text, extracting implicit schema information that aligns with formal conceptual models. These systems can generate candidate entity-relationship diagrams from document collections, identify potential cardinality constraints, and suggest derived attributes—all of which directly support CMML implementation. Current work is considering conceptual modeling and GenAI [77]. In addition, process modeling might help manage aspects of the overall ML process, from problem identification to deployment.

Applying conceptual modeling to ML can integrate or complement these two communities to support new interdisciplinary connections [54]. Our work complements these theoretical foundations by providing specific operational guidelines for incorporating conceptual modeling principles into ML data preparation activities. Our work is also intended to provide practical guidelines and contribute to the adoption of machine learning by providing some elements missing in AutoML [49].

## 5.2 Future Work

Building on the CMML method, we identify several promising research directions for our community. First, develop frameworks where conceptual models both inform machine learning processes and evolve based on patterns discovered through ML. Second, develop extensions to current modeling formalisms that better represent ML-specific concepts, such as semantic expressiveness, ontological commitments, and representational completeness. Third, develop automated techniques for extracting conceptual models from existing databases, enabling organizations without formal models to benefit from the performance improvements demonstrated in our empirical studies. Fourth, explore how conceptual models can enhance algorithmic transparency and fairness by formally representing sensitive attributes and domain constraints. Fifth, effectively incorporate CMML and possibly additional methods based on conceptual modeling into the routines used to perform automatic machine learning. These directions position conceptual modeling as a foundational element for responsible AI development. As our foster care applications demonstrate, this integration delivers measurable improvements in societally critical domains.

## 6 Conclusion

This research developed the Conceptual Modeling for Machine Learning (CMML) method comprised of guidelines for applying conceptual modeling concepts to the input data used for machine learning. The method was applied to two real-world cases in foster care management to show how it can improve model performance. CMML was further evaluated to examine its effects on data preparation process transparency using focus groups in which data scientists reviewed the guidelines and assessed their potential usefulness. Collectively, the evidence demonstrates that the method can improve important machine learning outcomes. The CMML method contributes to realizing the full potential of machine learning by supporting organizations in making and justifying data-driven decisions. It also contributes to the need to create socially responsible and effective machine learning initiatives. Since machine learning is increasingly being used in societally important domains, it needs to be deployed transparent, repeatable, and auditable, which can be supported by conceptual modeling. Finally, the research is an attempt to strengthen work between conceptual modeling and machine learning, which is part of the broader artificial intelligence community.

**Appendix A: Focus Group Protocol**

1. **Welcome (5 minutes).**
   Greet participants and allow each participant and researchers on call to briefly introduce themselves.

2. **Describe focus group procedures (5-10 minutes).**
The main difference between a group and a focus group is that a focus group has a specific, focused discussion **topic** and a trained **facilitator**. The group's composition is made up of a representative sample of individuals that might use our guidelines. We actively encourage you to
express your opinions but stay as much as possible on topic since we have a limited time window.

3. **Describe the objectives of the study.**
Machine learning faces many challenges to successful implementation and use. This research proposes how some of these challenges can be addressed by applying conceptual modeling concepts to data sets before using the data to build machine learning models.
We have proposed a series of guidelines that you received with your invitation. The guidelines are centered around the use of extended entity relationship (EER) diagrams to bring domain semantics into consideration and to provide a systematic approach for data preparation. We explained them with a running example.

These guidelines are based on conceptual modeling and their goal is to that ensure that data used in machine learning algorithms adheres to known domain semantics.

What do we mean by domain semantics (have slides ready with examples)? Sometimes, when we prepare data for ML models, we may miss important information about the underlying structure of the data which can lead to incorrect assumptions about how we should clean the data or interpret the results. This is very important in business data that is often transactional in nature (provide example of one to many – have slide prepared).

    a) Additionally, when we apply approaches such feature engineering, we often struggle to understand the role that our input data played in our predictions (provide example)

    b) Often, we have a hard time retracing our steps because data cleaning was not systematic.

Our goal is to find evidence of the utility and efficacy of these guidelines.

**4. Introduce CM constructs and guidelines (15-20 minutes):**

<u>Initial Impressions of the guidelines (10-15) minutes</u>

Imagine that you are building a model to predict customer lifetime value (have ERD up). Customer LTV is calculated from their activities (placing orders). The guidelines should help in organizing your data into a file that you would feed into a ML algorithm such as a Regression in R, Python, Rapid Miner.

Is there any guideline that you need me to explain?

You had a chance to look at the guidelines.

1. What do you think?
2. Which ones did you find useful for preparing data for ML?
3. Did any guidelines stand out?
4. Were any guidelines confusing?

<u>Cover any guidelines that were not mentioned</u>

1. Do you see yourself using these guidelines in the future?
2. What value does conceptual modeling play in our guidelines?
3. How do you think using these guidelines will affect the final outcome (performance? Interpretability/explainability?)